\documentclass[aps,pra,showpacs,preprintnumbers,amsmath,amssymb]{article}
\usepackage{arxiv}
\usepackage{mathptmx}
\usepackage{amssymb}
\usepackage{cite}  
\usepackage[pdftex]{graphicx} 
\usepackage{wrapfig} 
\usepackage{titlesec}
\usepackage{url}
\graphicspath{ {./images/} }
\usepackage{times}
\usepackage[overload]{textcase}
\normalfont

\title{
Speeding up reinforcement learning by 
combining attention and agency features
}

\author{	
Berkay Demirel\\
University Pompeu Fabra\\
Barcelona, Spain	 
\And 
Mart\'i S\'anchez-Fibla\thanks{Corresponding Author: Technology Department, Universitat Pompeu Fabra, Carrer de Roc Boronat 138, 08018 Barcelona, Spain. E-mail:
marti.sanchez@upf.edu.}\\
University Pompeu Fabra\\
Barcelona, Spain	 
}

\renewcommand{\headeright}{arXiv ver.}
\renewcommand{\undertitle}{arXiv version\thanks{Extended version of the paper presented in the International Conference of the Catalan Association for Artificial Intelligence, CCIA 2019 and appeared in 
		Artificial Intelligence Research and Development: Proceedings of the 22nd International Conference of the Catalan Association for Artificial Intelligence, Volume 319, 84, IOS Press, 2019.}}

\begin{document}
\maketitle

\begin{abstract}
When playing video-games we immediately detect which entity we control and we center the attention towards it to focus the learning and reduce its dimensionality. Reinforcement Learning (RL) has been able to deal with big state spaces, including states derived from pixel images in Atari games, but the learning is slow, depends on the brute force mapping from the global state to the action values (Q-function), thus its performance is severely affected by the dimensionality of the state and cannot be transferred to other games or other parts of the same game. We propose different transformations of the input state that combine attention and agency detection mechanisms which both have been addressed separately in RL but not together to our knowledge. We propose and benchmark different architectures including both global and local agency centered versions of the state and also including summaries of the surroundings. Results suggest that even a redundant global-local state network can learn faster than the global alone. 
Summarized versions of the state look promising to achieve input-size independence learning.
\end{abstract}


\section{Introduction}

When playing a video-game for the first time, we can quickly discern which entities are we in control (and when we are in control of those entities) by pressing buttons randomly, and we can direct attention to the immediate surroundings of those entities to reduce the state space dimensionality of the learning. Even in games with first person view we constantly direct attention towards what we can have an effect on.
Decomposing the perceived state into different entities is fundamental for generalization and transfer learning \cite{burgess2019monet}. Attentional mechanisms are crucial in the brain to direct learning abilities \cite{niv2015reinforcement}. Reinforcement Learning (RL) has been able to deal with big state spaces, including states derived from pixel level images (i.e. Deep Q Network, DQN, learning to play Atari games \cite{mnih2015human}) thanks to the usage of Deep Convolutional Neural Networks to approximate the mapping of states and actions to future reward. 
The performance of RL algorithms like DQN is severely affected by the size/dimensionality of the input state spaces, which does not seem to be the case for human play performance. We believe that this input size independence may be achieved by the combination of visual attention \cite{mnih2014recurrent} and agency detection mechanisms \cite{thomas2017independently,brody2017learning,sanchez2017social} to deliver an agent-centered view of the state to the RL algorithm. 
We argue and prove experimentally that it is beneficial to center the state where the agent can cause immediate effects. Thus we are proposing to deliver a first person view of the state centered where the actions cause effects and where the agent has direct agency. This agency-centered state can have higher resolution in the center and lower in the surroundings and can include a summary of the important features that are out of reach. 
Some computer games (like Doom, Quake, Obstacle Tower Challenge) or robotics applications (like the one we describe in \cite{sanchez2019motor}) are naturally first person view (or close to it) and we think that they can still benefit from our approach as a first person view can be centered towards the focus of agency, the part of the view in which the agent can have the maximum effect. In first person shooter games like Doom, the target of the gun could be a possible candidate; in a robotics hand object manipulation this would be the case of the hand.     

We propose and benchmark different architectures including both global and local agency centered versions of the state and also including summaries of the surroundings. Results suggest that even a redundant global local state network can learn faster than the global alone.

\section{Methods}
\label{methods}

We model the environment by a Markov Decision Process (MDP). An MDP is a tuple $\langle S,A,P,R, \gamma \rangle$ where $S$ is the state of states, $A$ the set of actions, $P$ the transition probability distribution $P: S \times A \times S \rightarrow [0,1]$ satisfying $\sum_{s'}P(s'|s,a)=1$, $R$ the expected reward function for every state and action $R: S \times A \rightarrow \mathbb{R}$ and $\gamma$ the discount factor of the return (reward obtained from state $s$ and discounted geometrically by $gamma$): $G_t = R_{t+1} + \gamma R_{t+2} + \gamma^2 R_{t+3} + ...$. The aim of Reinforcement Learning (RL) is to learn an optimal policy that maps states to actions $\pi: S \rightarrow A$ such that future expected reward is maximized:  $\mathbb{E}_{\pi}[G_t | S_t=s]$ being $S_t$ the state at time $t$.

RL algorithms like Q-Learning solve the problem by estimating the so-called action-value function $Q: S \times A \rightarrow \mathbb{R}$ from which an optimal policy can be obtained by "greedyfication": $\pi^*(s) = \mathrm{argmax}_a Q(s,a)$

When the state space is high dimensional as in the case of Atari games where a state $s \in \mathbb{R}^{210x160x3}$ a Deep Convolutional Neural Network (DCNN) is used to approximate $Q$ as in the case of DQN algorithm in which also a technique named experience replay is used to update the DQN function in batches of stored experience so to make the samples as much as independent and identically distributed (i.i.d) as possible.

\subsection{Setup}
\label{setup}

We use a modified version of the Fruit Collection environment \cite{seijen2017environment}. It consists of an 11 by 11 grid world, the outer most layer consisting of walls, effectively limiting the positions where the agent can move to 81. 
The agent as well as 5 fruits are initialized to randomly distinct positions on the grid. 
The agent has 4 available actions to move in the 4 cardinal directions: up, down, left and right.
The environment used in "pixel mode", has what we call a global state  $s^{global} \in \mathbb{R}^{11x11x3}$, 3 channels sized 11 x 11 corresponding to agent location, fruit locations and walls respectively. 
Each time the agent enters a position where a fruit is present, it is rewarded with +1 and the fruit disappears. An episode ends when all fruits are collected or after 80 steps, whichever comes first.
This pixel mode allows the use of convolutions as it keeps the spatial relationship between features, while keeping the training required for feature recognition to a minimum by one hot encoding of features on different layers. 

\subsection{Agency Detection}

From the described global state $s^{global} \in \mathbb{R}^{11x11x3}$ one could compute the coordinates of the agent that is controlled by the available actions. There exist different studies that deal with the problem of identifying controllable features of the state \cite{thomas2017independently}. In \cite{brody2017learning} this is called learning agency.
In \cite{sanchez2017social} the controllable features of the state are assessed by computing correlations of the motor actions into the optical flow effects of the different visual features. In the following we are going to assume that we know the coordinates in the global state of the agent that is controlled by the available actions and that is what we mean when we say we assume knowledge of agency.

\subsection{State Transformations}
\label{state}

\begin{figure}[htbp]
  \begin{center}
    \includegraphics[clip=True,width=\textwidth]{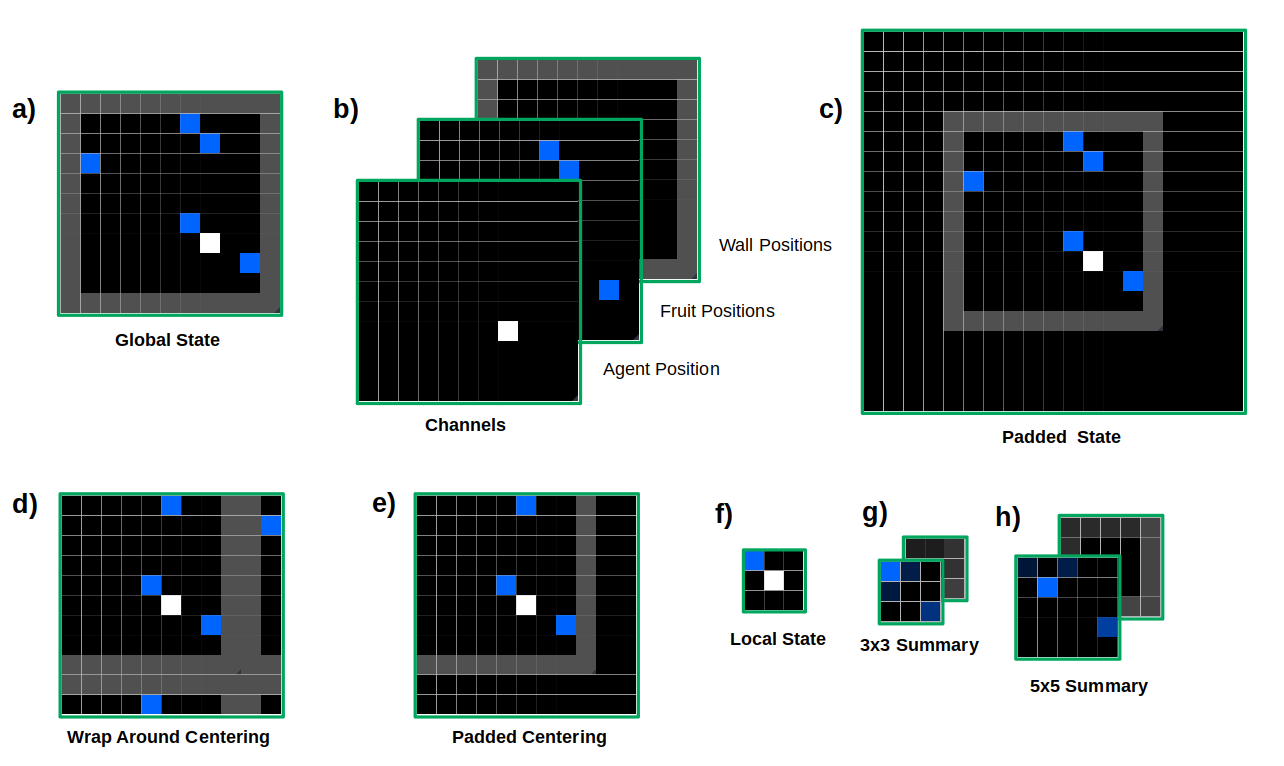}
  \end{center}
  \caption{State Transformations. \textbf{a),b)} Global state  $s^{global} \in \mathbb{R}^{11x11x3}$ and its different 3 channels: walls (grey squares), fruits (in blue) and the agent position (white square). \textbf{c)} padded state result of  $f^{pad}_{16x16}(s)$ operation in which all channels are filled with zeros in the surroundings. \textbf{d)} Centering with wrap around result of $f^{center}_{roll}(s) = f^{roll}_{x,y}(s)$ operation. \textbf{e)} Centering operation after padding $f^{center}_{pad}(s) = f^{sub}_{11x11}(f^{roll}_{x,y}(f^{pad}_{16x16}(s)))$. Some information might be lost and the state becomes partially observable. \textbf{f)} Local state extracted from the centered state version: $s^{local}_{pad} = f^{sub}_{3x3}(f^{center}_{pad}(s))$. \textbf{g),h)} Summarized versions of the local state: $s^{summary}_{3x3}$ and $s^{summary}_{5x5}$}
  \label{transforms}
\end{figure}

We describe in the following the different operations that we use to transform the global state into an agent-centered view and reduced local view around the agent (see Figure \ref{transforms}).

The first operation, that we call centering, enables to center the state at a certain position $i,j$ of the matrix (this position will be the one of the agent). We consider two possible centering modes: rolling $f^{center}_{roll}$ and padding $f^{center}_{pad}$. 
When centering by rolling the roll operation is used $f^{roll}_{x,y}$ which does a circular rolling of the rows (by $x$ positions) and columns (by $y$ positions) without losing information (as if the environment had a torus topology in which the extremes are connected). Centering to position $i,j$ can be achieved by rolling $(nrows+1)/2-i$ the rows of each channel and rolling $(ncols+1)/2-j$ the columns of each channel (being $nrows$ and $ncols$ the number of rows and columns respectively of the matrix).  As  $s^{global} \in \mathbb{R}^{11x11x3}$ centering by rolling is: $f^{center}_{roll}(s) = f^{roll}_{6-i,6-j}(s)$.

Centering can also be achieved with padding: each matrix is padded with zeros adding rows and columns, $f^{center}_{pad}(s) = f^{sub}_{11x11}(f^{roll}_{i,j}(f^{pad}_{16x16}(s)))$. In the case of centering with padding some information might be lost when the agent is near the limits of the environment.

Once centered a sub-matrix can be extracted so as to discard the outer rows and cols of the matrix thus extracting what we call the local state, considering the immediate surroundings of the agent that is controlled by the actions. For example the local agent centered view with padding would be: $s^{local} = f^{sub}_{3x3}(f^{center}_{pad}(s))$.

Finally we describe the "summary transformations" of the state. 
Consider the padded centered state as made up of progressively larger layers surrounding the agent like an onion. We eliminate iteratively each external layer by taking the mean of the weighted sum into the inner layers. This is realized in multiple steps. First, the value of each point in the outermost layer of the state matrix is multiplied by a constant of 0.9. This makes sure that the outer layers are weighted less as they are far from the immediate reach of the agent. Secondly, the mean of each point and it's two immediate neighbours is taken and added to the value of the point that is directly one layer below it in the state matrix. This operation is repeated until a 3x3 matrix is obtained in the case of $s^{summary}_{3x3} = f^{summary}_{3x3}(f^{center}_{pad}(s))$ or stopping earlier and having the summary values surround the local state space to create a 5x5 matrix:
$s^{summary}_{5x5} = f^{summary}_{5x5}(f^{center}_{pad}(s))$.
The "summary transformations" have the nice property of being local (while preserving a weak form of total observability) and independent of the size of the environment. "Summary transformations" also depend heavily on the existence of a one-hot encoded feature map, similar to our state representations as the RGB colour values could not be meaningfully summarized with this approach.

The final four important state transformations that will be used in the following for the learning architectures are:
\begin{itemize}
    \item Global state: the intact $s^{global} \in \mathbb{R}^{11x11x3}$ (see Figure \ref{transforms}a).
    \item Global centered-roll state: $f^{center}_{roll}(s)$ (See Figure \ref{transforms}d). 
    \item Global centered-padded state: $f^{center}_{pad}(s)$ (See Figure \ref{transforms}e).
    \item Local state:  $s^{local} = f^{sub}_{3x3}(f^{center}_{pad}(s))$ (See Figure \ref{transforms}f).
    \item Summary states: $s^{summary}_{3x3}$  and $s^{summary}_{5x5}$ (See Figure \ref{transforms}g and  \ref{transforms}h).
\end{itemize}

\subsection{Agent and Architectures}

The agent implements a state of the art reinforcement learning algorithm that in our case we have selected Deep Q Learning, DQN  \cite{mnih2015human} with prioritized experience replay. As mentioned previously, in DQN, the Q-function is approximated using neural networks.
The network consists of an input layer corresponding to each variation's global state size: 3x11x11 for the no centering architectures and 2x11x11 for both centering with padding and centering with wrap around architectures. The first hidden layer is a convolutional layer consisting of 12 kernels of size 3x3, with a stride of 1 and rectified linear unit (relu) activation function. The second hidden layer is another convolutional layer of 16 kernels of size 3x3, with a stride of 1 and relu activation function. On the third layer, the output of the last layer is flattened before passing densely to 16 nodes with a relu activation function. The output layer consists of 4 nodes, each corresponding to a directional movement action. Huber loss is used as the loss function and Adam with a learning rate of 0.0002 as the optimizer. 

The target network weights are updated every step using Polyak averaging with a tau of 0.0005. Each agent is trained for 800 episodes with a discount factor of 0.95. Epsilon decays linearly from 1 at the $1^{st}$ episode to the final value of 0.05 at $750^{th}$ episode. The experience replay buffer size is 10000 for the architectures that use the global state, while it is 2400 for the architectures that utilize the local state. The training starts after the buffer is filled by a random policy agent. At each step a batch of size 32 is selected from the buffer with a chance proportional to the size of the TD-error with a guarantee that every experience will be replayed at least once. The weights of the experiences are inversely proportional to their TD-error to offset the selection bias, and updated for each experience whenever they are replayed.

\subsection{Global Architecture}

This architecture consists of 3 variations:
\begin{description}
  \item[$\bullet$] No centering variation assumes no agency knowledge and as such the 3x11x11 state returned from the environment is directly passed on to the agent without any transformations.

  \item[$\bullet$] Centering with padding variation assumes agency knowledge. This variation pads the environment with 0s before centering it around the agent. The layer representing the agent position is not required because of the agent-centered view and is dropped. The agent is passed a 2x11x11 state. This variation also turns the problem from a Markov Decision Process to Partially Observable Markov Decision Process as the Markov property does hold because of the loss of information in the transformation.

  \item[$\bullet$] Centering with wrap around variation assumes agency knowledge. This variations wraps the environment around itself from both axes. The layer representing the agent position is not required because of the agent-centered view and is dropped. The agent is passed a 2x11x11 state. This variation does not turn the problem into a POMDP as the Markov property still holds because there is no information loss in the transformation.

\end{description}

\begin{figure}[htbp]
  \begin{center}
    \includegraphics[clip=True,width=0.85\textwidth]{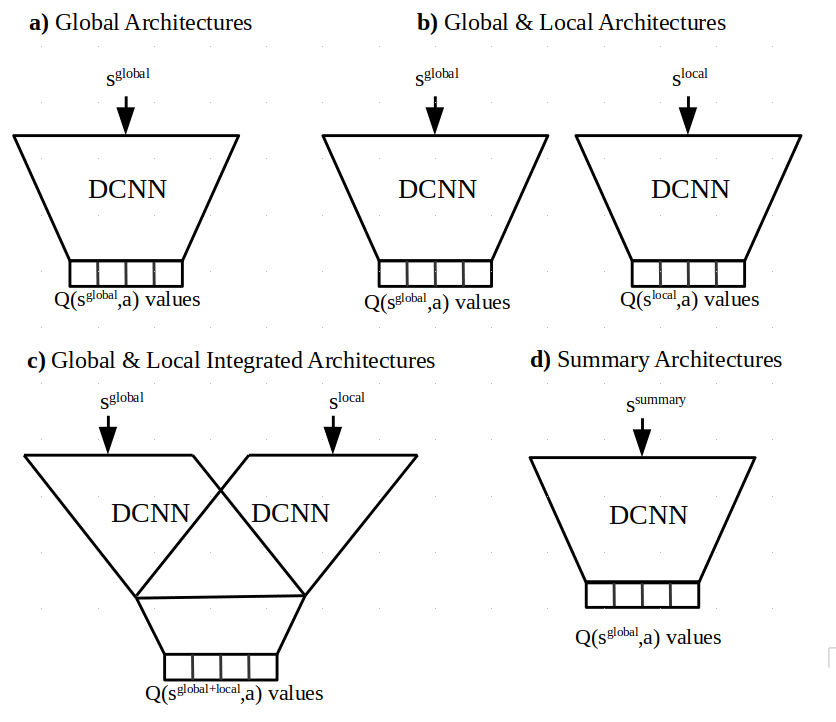}
  \end{center}
  \caption{Learning Architectures. \textbf{a)} Global architecture which takes as input the global state $s^{global}$. The network is a Deep Convolutional Neural Network (DCNN). \textbf{b)} Two separate networks have as input $s^{local}$ and $s^{global}$ respectively. \textbf{c)} A single network receives as input both $s^{local}$ and $s^{global}$. \textbf{d)} A single network receives the summarized version of the state $s^{summary}$.}
  \label{archs}
\end{figure}

\subsection{Global and Local Architecture}

In this architecture, the global state is decomposed into 2 parts: one local and one global. While the global one is the same as the one used in the global architecture, the local state is the 3x3 grid surrounding the agent denoted as $s^{local}_{pad}$ in section \ref{state}. The agent consists of 2 different networks: the global network and the local network, taking as input $s^{local}$ and $s^{global}$ respectively. Action selection is mediated by a simple conditional rule: if a feature exists in the local state, the action is selected according to the local network, otherwise, the global network is used for action selection. This rule is selected for it's simplicity, effectively making the agent rely on the global network for navigation while using the local network for reacting to the local stimuli. As it does not utilize the global state when there is a feature in the local space, it is not expected to be optimal but rather serve as a test for the performance of the agent when different networks for different state spaces are used. More complex action selection schemes such as a confidence measures can also be utilized.

\begin{description}
  \item[$\bullet$] No centering: This variation is similar to the same variation of the global architecture with the addition of a decomposed local state centered around the agent. The global state is a matrix of size 3x11x11, while the local state is 2x3x3. 

  \item[$\bullet$] Centering with padding: The global state is centered around the agent, the agent position channel dropped and the necessary parts padded with 0s. The global state is a matrix of size 2x11x11, while the local state is 2x3x3. 

  \item[$\bullet$] Centering with wrapping around: The environment is wrapped around itself from both axes, the global state is centered around the agent and the agent position channel dropped. The global state is a matrix of size 2x11x11, while the local state is 2x3x3.
\end{description}

\subsection{Global and Local Integrated Architecture}

This architecture differs from the previous Global and Local network by getting rid of the rule for network selection and integrating both $s^{local}$ and $s^{global}$ into the same network in a decomposed manner. The global network is modified by adding a secondary input layer that takes the local 3 by 3 state around the agent. This local state $s^{local}$ is convolved by 6 kernels with size 2x2 with a single stride and flattened afterwards. This flattened local state is concatenated with the flattened global state before connecting to the dense layer of 16 nodes and passed to the output layer. This allows the network to learn when to rely on the local state, as the local state space would be empty when no features are detected and thus the output of the network would solely rely on the global state. Similar to the global and local architecture, this architecture has 3 variations involving how the global state is transformed before passing along to the agent:

\begin{description}
  \item[$\bullet$]  No centering variation involves passing the global state to the agent without any transformation. The global state is a matrix of size 3x11x11, while the local state is a matrix of size 2x3x3.

  \item[$\bullet$] Centering with padding involves passing the global state to the agent by centering the view on the agent, dropping the agent location channel and padding the necessary parts with 0s. As a result, the global state is a matrix of size 2x11x11, while the local state is a matrix of size 2x3x3. 

  \item[$\bullet$] Centering with wrapping around involves wrapping the global state on both axes around itself before centering it on the agent and dropping the agent location channel. The global state is a matrix of size 2x11x11, while the local state is of size 2x3x3.
\end{description}

\subsection{Summary Architecture}

This architecture utilizes the local summary state space transformation that we call $s^{summary}$ in section \ref{state}. The state is reduced to a lower dimensionality state by doing a weighted sum of the outer layers of the state towards the inner ones. The architectures takes an input a reduced version of the global state that maintains a weak total observability.

\begin{description}
  \item[$\bullet$]  3x3 Summary: In this variation the global environment is collapsed onto itself until a 3 by 3 grid surrounding the agent remains. 

  \item[$\bullet$]  5x5 Summary: In this variation the 3 by 3 layer surrounding the agent is kept, while the environment is collapsed around this local view, creating a 5 by 5 grid. 
\end{description}

\section{Results}
\label{results}

\begin{figure}[htbp]
\includegraphics[width=0.9\textwidth]{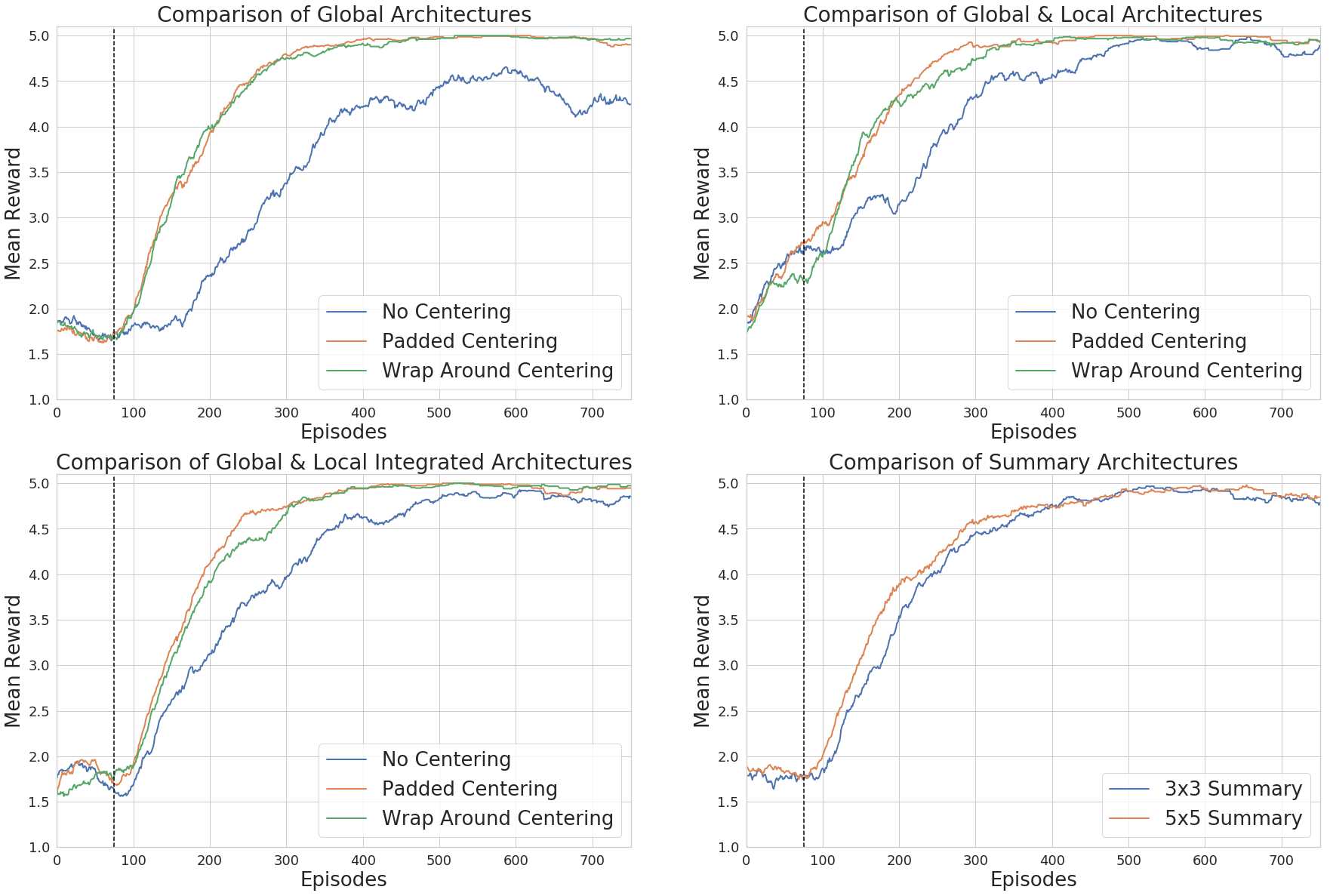}
\caption{Running mean rewards for all the variations}
\label{means}
\centering
\end{figure}

Figure \ref{means} shows the mean reward obtained by each variation and architectures (we use a running window average 50 episodes). In the first subplot, variations with centering show a much higher learning performance from the beginning and converge much earlier than the global no centering variation. This can be attributed not only to the more compact state space representation by getting rid of the agent location channel, but also to the location invariance resulting from centering the state. The compact state space representation is more informative and as a result less examples and training time is needed for convergence. There is no difference in the learning performance of the padded centering and wrap around centering variations. Although this lack of difference in performance is unintuitive given that the padded centering variation is partially observable, it can be attributed to the simplicity of the environment and the task. 

\begin{figure}[b!]
  \begin{center}
    \includegraphics[clip=True,width=0.7\textwidth]{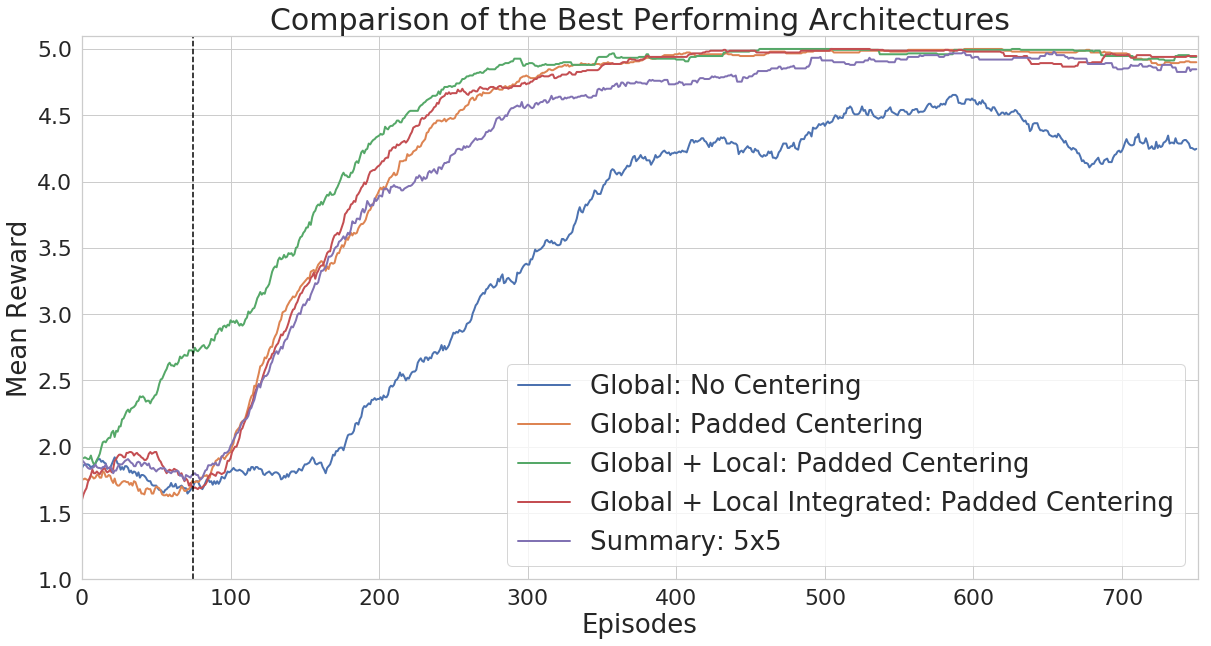}
  \end{center}
  \caption{Comparison of the best performing architectures}
  \label{bestperform}
\end{figure}

The results of the global and local architecture are similar to the global architecture, showing higher performance for the centering variations compared to the no centering variation. The global and local integrated architecture, similar to all other architectures show increased performance of centered variations, with a lower and less stable learning performance for the no centering variation.

The results of the summary architecture show that the 5x5 summary variation has a small but significant learning advantage compared to the 3x3 summary variation. Although the state representation is larger in the 5x5 variation, the advantage afforded by keeping the immediately surrounding layer around the agent offsets this to such a degree that it creates an advantage. 

A comparison of the best performing architectures can be seen in Figure  \ref{bestperform}. The best and the worst performing architectures are global and local with padded centering and global without centering respectively. The other 3 architectures are clustered more closely together, with the 5 by 5 summary following behind the padded centering variation of the global architecture and the global and local integrated architecture with padded centering.

\begin{figure}[htbp]
  \begin{center}
    \includegraphics[clip=True,width=0.55\textwidth]{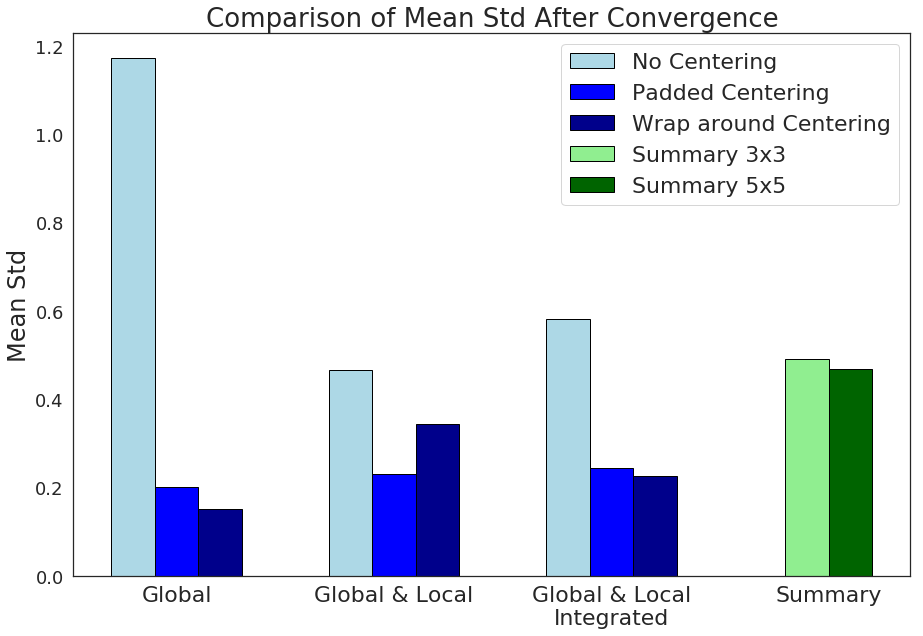}
  \end{center}
  \caption{Mean standard deviation comparison after convergence}
  \label{stability}
\end{figure}

The end of training stability measured as the mean of standard deviation after the convergence of most architectures, which is around the $500^{th}$ episode mark, is shown in Figure \ref{stability}. It can be seen that all the architectures with centering are more stable compared to their no centering counterparts. This difference is most pronounced in the global architecture. The results do not show a clear stability advantage for either of the centering methods. The summary architectures performed similarly, with the 5x5 variation showing a tiny advantage over the smaller counterpart.

\section{Discussion}
\label{discussion}

The results show that using agency information to center the state space around the agent increases performance regardless of the architecture.

The global and local architecture's centering variations show the highest learning performance among all the architectures. This is especially pronounced in early learning, as the local network converges much faster as a result of much smaller local state space. The conditional action selection rule forces the agent to act close to optimality from very early on. This can be seen both as an advantage and as a disadvantage, as the agent becomes reactive very quickly. This means that whenever there is a reward in a single-step distance, the agent learns to prioritize picking it up, but this can result in sub-optimal behavior in a wide variety of cases. For example, the agent would prefer to pick up a single reward one step away instead of the 4 rewards that are 2 steps away. This blindness to outside it's immediate surroundings causes it to perform well over what a global agent does at the beginning but towards the end of the training it causes instability as it continues to act sub-optimally from time to time. The global agent, on the other hand, takes longer to learn but considers the whole state before acting thus acts consistently optimal after a certain point. The global and local integrated architecture performs well above the global architecture even when there is no centering. 

The simultaneous usage of global and local versions of the state in the proposed architectures reminds of nested reactive and contextual controllers as in the Distributed Adaptive Control \cite{maffei2015embodied} (see for example different ways of combining reactive and adaptive control via reinforcement learning \cite{freire2018modeling,moulin2015autonomous}). The difference here is that even the reactive is learned through interaction with the environment. The local state can be used for reactive actions and the global state for long term planning.

\section{Conclusion}
\label{conclusion}

We have proposed different transformations of the input state of a Reinforcement Learning (RL) agent that combine attention and agency detection features; both have been addressed separately in RL but not jointly to our knowledge. We have proposed and benchmarked different architectures including both global and local agency centered versions of the state and also including "summaries" of the surroundings. Results suggest that even a redundant global local state network can learn faster than the global alone. "Summarized" versions of the state look promising to achieve input-size independence learning.

Future studies will explore how the state transformations proposed in this paper scale to larger state spaces with higher dimensions, and whether the promise of "summary transformations" to achieve input-size independence holds. 
We will also investigate alternative architectures that can utilize progressively larger state spaces as the training progresses.

\section*{Acknowledgments}
Research supported by INSOCO-DPI2016-80116-P.

\bibliographystyle{plain}
\bibliography{agency-attention}

\end{document}